%% file: ms.tex
\documentclass[10pt,twocolumn,letterpaper]{article}

\usepackage{cvpr}
\usepackage{times}
\usepackage{epsfig}
\usepackage{graphicx}
\usepackage{amsmath}
\usepackage{amssymb}
\usepackage{xcolor}
\usepackage{float}
\usepackage{cuted}
\usepackage{capt-of}

\usepackage{blindtext} 
\usepackage{lipsum} 



\usepackage[breaklinks=true,bookmarks=false]{hyperref}

\newcommand{\calG}{\mathcal{G}}
\newcommand{\calE}{\mathcal{E}}
\newcommand{\calV}{\mathcal{V}}
\newcommand{\calT}{\mathcal{T}}
\newcommand{\calD}{\mathcal{D}}
\newcommand{\calZ}{\mathcal{Z}}

\newcommand{\bfd}{\mathbf{d}}

\newcommand{\rmT}{\mathrm{T}}

\newcommand{\bfX}{\mathbf{X}}

\newcommand{\homod}{\tilde{\bfd}}
\cvprfinalcopy 

\def\cvprPaperID{5122} 

\begin{document}

\title{4D Association Graph for Realtime Multi-person Motion Capture \\ Using Multiple Video Cameras}

\author{Yuxiang Zhang$^{1}\thanks{Equal contribution}$, Liang An$^{1*}$, Tao Yu$^1$, Xiu Li$^1$, Kun Li$^2$, Yebin Liu$^1$\\
$^1$Institute for Brain and Cognitive Sciences, Tsinghua University, China \ \ $^2$Tianjin University, China\\
}

\maketitle

\input{version2/0_abstract.tex}

\input{version2/1_introduction.tex}
\input{version2/2_related_work.tex}
\input{version2/3_overview.tex}
\input{version2/4_modeling.tex}
\input{version2/5_solving.tex}
\input{version2/7_results.tex}
\input{version2/8_diss.tex} 

{\small
\bibliographystyle{ieee_fullname}
\bibliography{egbib}
}

\end{document}

%% file: version2/0_abstract.tex
\begin{abstract}
This paper contributes a novel realtime multi-person motion capture algorithm using multiview video inputs. Due to the heavy occlusions in each view, joint optimization on the multiview images and multiple temporal frames is indispensable, which brings up the essential challenge of realtime efficiency. To this end, for the first time, we unify per-view parsing, cross-view matching, and temporal tracking into a single optimization framework, i.e., a 4D association graph that each dimension (image space, viewpoint and time) can be treated equally and simultaneously. To solve the 4D association graph efficiently, we further contribute the idea of 4D limb bundle parsing based on heuristic searching, followed with limb bundle assembling by proposing a bundle Kruskal's algorithm. Our method enables a realtime online motion capture system running at 30fps using 5 cameras on a 5-person scene. Benefiting from the unified parsing, matching and tracking constraints, our method is robust to noisy detection, and achieves high-quality online pose reconstruction quality. The proposed method outperforms the state-of-the-art method quantitatively without using high-level appearance information. We also contribute a multiview video dataset synchronized with a marker-based motion capture system for scientific evaluation. 
\end{abstract} 

%% file: version2/1_introduction.tex
\vspace{-2mm}

\section{Introduction}
Markerless motion capture of multi-person in a scene is important for many industry applications but still challenging and far from being solved. Although the literatures have reported single view 2D and 3D pose estimation methods~\cite{wei2016convolutional,pishchulin2016deepcut,cao2018openpose,Chen2018CPN,he2017mask,fang2017rmpe,li2018crowdpose,nie2019single,zanfir2018monocular,zanfir2018deep,singleshotmultiperson2018}, they suffer from heavy occlusions and produce low-fidelity results. Comparably, multi-view cameras provide more than one views to alleviate occlusion, as well as stereo cues for accurate 3D triangulation, therefore are indispensable inputs for markerless motion capture of multi-person scenes. While remarkable advances have been made in many kinds of multi-camera motion capture systems for human \cite{liu2013markerless,liu2011markerless,joo2019panoptic} or even animals \cite{bala2020openmonkeystudio}, most of them fail to achieve the goals of realtime performance and high quality capture under extremely close interactions. 


\begin{figure}
    \centering
    \includegraphics[width=0.95\linewidth]{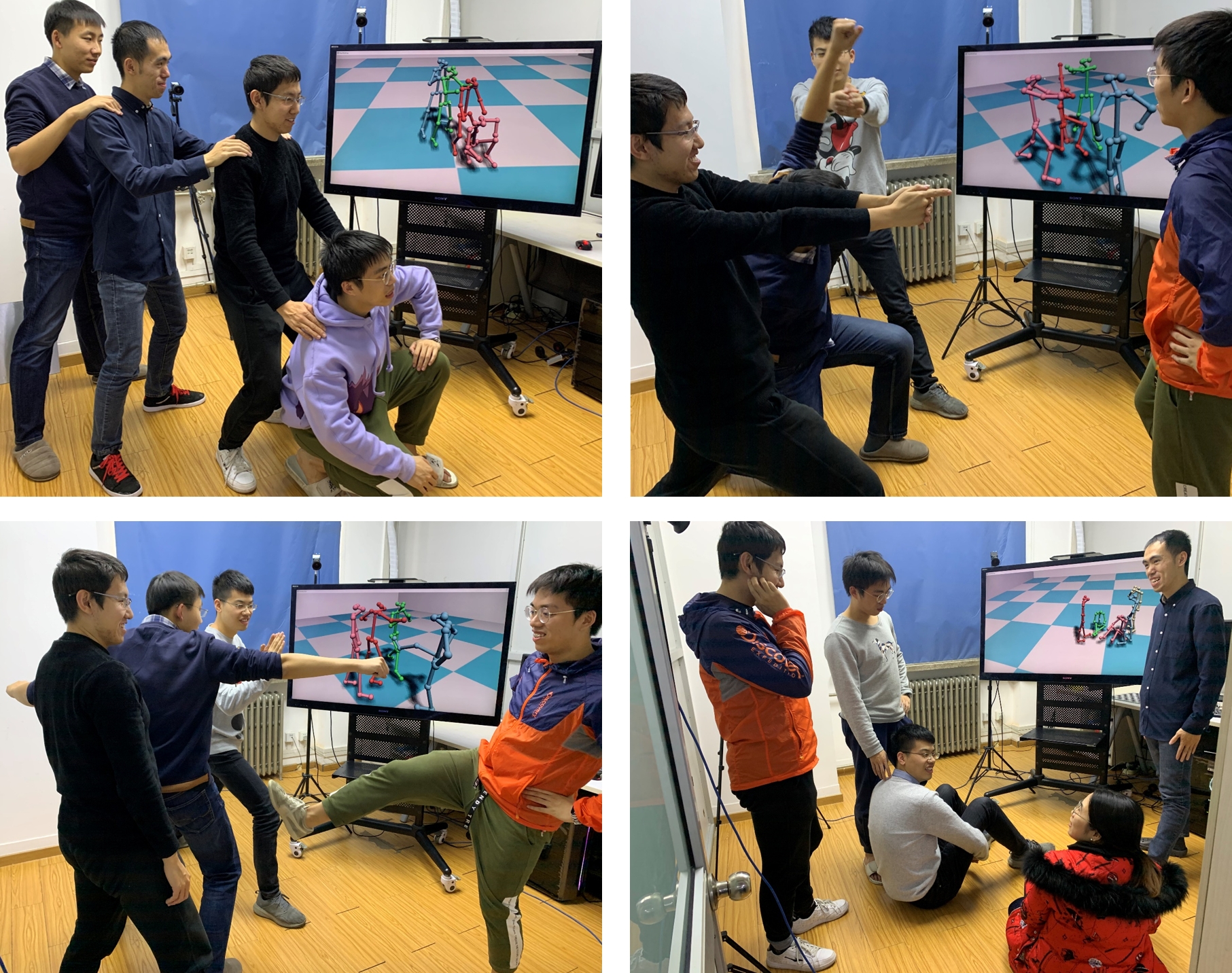}
    \caption{Our method enables multi-person motion capture system working at 30fps for 5 persons using 5 RGB cameras, while achieving high quality skeleton reconstruction results.}
    \label{fig:teaser}
     \vspace{-2mm}
\end{figure}


Given the 4D (2D spatial, 1D viewpoint and 1D temporal) multiview video input, the key to the success of realtime and high quality multi-person motion capture is how to leverage the rich data input, \textit{i.e.}, how to operate on the 4D data structure to achieve high accuracy while maintaining realtime performance. Essentially, based on the human body part features pre-detected in the separate 2D views using state-of-the-art CNN methods~\cite{cao2018openpose}, three kinds of basic associations can be defined on this 4D structure. These include single image association ({\textit{i.e.}, \bf parsing})~\cite{cao2018openpose,deepercut} to form human skeletons in a single image, cross-view association ({\textit{i.e.}, \bf matching}) to establish correspondences among different views, and temporal association (\textit{i.e.} {\bf tracking}) to build correspondences between sequential frames.

Existing methods struggle to deal with all these association simultaneously and efficiently. They consider only parts of these associations, or simply operate them in a sequential manner, resulting in failure to be a high quality and realtime method. For example, the state-of-the-art methods \cite{dong2019fast,bridgeman2019multi,tanke2019iterative} share a similar high-level framework by first performing per-view person parsing, followed by cross-view person matching, and temporal tracking sequentially. They usually assume and rely on perfect per-view person parsing results in the first stage. However, this can not be guaranteed in crowded or close interaction scenarios. Temporal extension \cite{belagiannis2014temporal3DPS,belagiannis20163dps} of the 3D pictorial structure (3DPS) model \cite{belagiannis20143d} apply temporal tracking \cite{multi-object-tracking}, followed with cross-view parsing using the very time-consuming 3DPS structure optimization. The Panoptic Studio \cite{joo2019panoptic} addresses these associations in a sequential manner, by first matching (generate node proposals), then tracking (generate trajectories), and finally assemble the 3D human instances. As it tracks over the whole sequence, it is impossible to achieve realtime performance. 


In this paper, we formulate parsing, matching, and tracking in a unified graph optimization framework, called {\bf 4D association graph}, to simultaneously and equally addressing 2D spatial, 1D viewpoint and 1D temporal information. By regarding the detected 2D skeleton joint candidates in the current frame and the 3D skeleton joints in the former frame as graph nodes, we construct edges by calculating confidence weights between nodes. Such calculation jointly takes advantage of feature confidences in each individual image, epipolar constraints and reconstructed skeletons in the temporal precedent frame. Compared with \cite{dong2019fast,joo2019panoptic,belagiannis2014temporal3DPS,belagiannis20163dps} which adopt sequential processing strategy on image space, viewpoint, and time dimensions, our 4D graph formulation enables unified optimization on all these dimensions, thereby allowing better mutual benefit among them. 

To realize realtime optimization on the 4D association graph, we further contribute an efficient method to solve the 4D association by separating the problem into a 4D limb parsing step and a skeleton assembling step. In the former step, we propose a heuristic searching algorithm to form 4D limb bundles and a modified minimum spanning tree algorithm to assemble the 4D limb bundles into skeletons. Both of these two steps are optimized based on an energy function designed to jointly consider the image feature, stereo and temporal cues, thus optimization quality is guaranteed while realtime efficiency is achieved. We demonstrate a realtime multi-person motion capture system using only 5 to 6 multiview video cameras, see Fig.~\ref{fig:teaser} and the supplemental video. Benefiting from this unified strategy, our system succeeds even in the close interaction scenarios (Video 02:55-03:30). Finally, we contribute a multiview multi-person close interacting motion dataset synchronized with marker-based motion capture system.

%% file: version2/2_related_work.tex
\section{Related Work}
We briefly overview literature on multi-person skeleton estimation according to the dimension of input data.
\subsection{Single Image Parsing}
We restrict our single image parsing to the work that addresses multi-person pose estimation in 2D and 3D. As there are close interactions in the scene, they all need to consider skeleton joint or body part detection and their connection to form skeletons. Parsing methods can be typically categorized into two classes: bottom-up method and top-down method. In general, top-down methods \cite{ke2018multi,fang2017rmpe,Chen2018CPN,he2017mask,xiao2018simple,li2018crowdpose} demonstrate higher average precision benefiting from human instance information, and bottom-up methods ~\cite{deepercut,cao2018openpose,papandreou2018personlab,kocabas2018multiposenet,Jie:graph:decomposition} tend to propose pixel-aligned low-level feature positions while assembling them is still a great challenge.
Typically, a state-of-the-art bottom-up method, OpenPose~\cite{cao2018openpose}, introduces part affinity field (PAF) to assist parsing low-level keypoints on limbs, obtaining realtime performance with high accuracy.

\subsection{Cross-view Matching}
Matching finds correspondences across views, no matter on high level features (human instances) or low-level features (keypoints). Previous work \cite{belagiannis20143d,belagiannis2014temporal3DPS,belagiannis20163dps,ershadi2018multiple} implicitly solves matching and parsing using 3D pictorial structure model. However, such method is time-consuming due to large state space and iterative belief propagation. Joo \emph{et al}. \cite{joo2019panoptic} utilize detected features from dense multi-view images to vote for possible 3D joint positions, which does matching in another implicit way. Such voting method only works well with enough observation views. Most recent work \cite{dong2019fast} matches per-view parsed human instances cross view with convex optimization method constrained by cycle-consistency. Though fast and robust, such method relies on appearance information to ensure good results, and could be affected by possible parsing error (\emph{e.g}. false positive human instance and wrong joint estimation).

\subsection{Temporal Tracking}
Tracking is one key step towards continuous and smooth motion capture, and helps solve current pose ambiguity according to history results. Tracking could be done either in 2D space or 3D space. Many works have addressed 2D tracking, known as pose-tracking tasks \cite{andriluka2018posetrack,CVPR2019_staf,iqbal2017posetrack,insafutdinov2017arttrack}. For 3D tracking, motion capture of multiple closely interacting persons \cite{liu2011markerless,liu2013markerless} has been proposed through joint 3D template tracking and multi-view body segmentation. Li \emph{et al}. \cite{li2018PG} propose a spatio-temporal tracking for closely interacting persons from multi-view videos. However, these pure tracking algorithms are easy to fail because of temporal error accumulation. Elhayek \emph{et al}.~\cite{elhayek2017marconi} track 3D articulated model to 2D human appearance descriptor (Sum of Gaussian), achieving markerless motion capture for both indoor and outdoor scenes. However, it does not demonstrate multi-person case (more than 3 persons). Belagiannis \emph{et al}.~\cite{belagiannis2014temporal3DPS} also utilize tracking information, but they derive human tracks in advance as prior to reduce state space, instead of solving tracking and matching simultaneously. Bridgeman \emph{et al}.~\cite{bridgeman2019multi} contribute a real time method, yet it adopt a sequential processing of image parsing, cross-view correction and temporal tracking, resulting in degraded accuracy compared with the state-of-the-art \cite{dong2019fast}. In Panoptic Studio \cite{joo2019panoptic}, after temporal tracking of 3D joint proposals on the whole sequence, optimization is started for human assembling.   

%% file: version2/3_overview.tex
\section{Overview}
\label{sec:formulation}
\begin{figure*}[ht!]
    \centering
    \includegraphics[width=\linewidth]{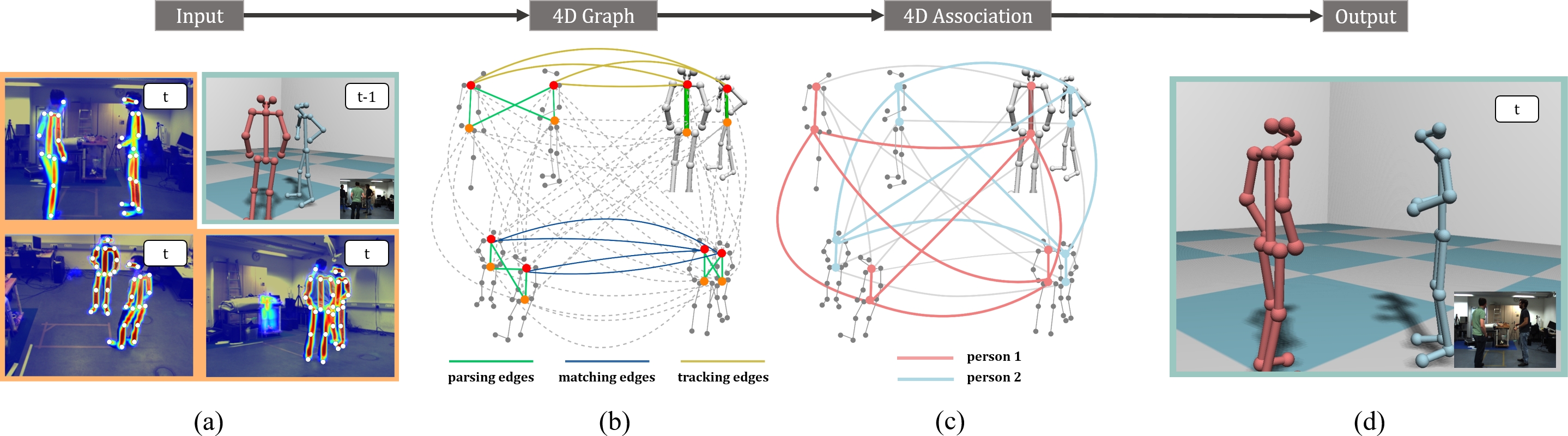}
    \vspace{-3mm}    
    \caption{Method overview. (a) We input body part positions and connection confidence of different views at time $t$, together with 3D person of last time. We use 3 views for example. (b) The 4D association graph. For clarity, we only highlight the association of the torso limb with three types of edges (\textbf{parsing} edges, \textbf{matching} edges and \textbf{tracking} edges) with different colors. (c) From the initial graph (b), our association method outputs the assembling results. (d) We optimize the assembled multiview 2D skeletons (c) to form 3D skeletons of current frame $t$.
    }
    \label{fig:overview_system}
    \vspace{-3mm}
\end{figure*}

Our 4D association graph considers the information in two consecutive frames. We first use the off-the-shelf bottom-up human pose detector \cite{cao2018openpose} on each input view of the current frame to generate low-level human features on each view. Our 4D association graph takes as input multi-view human body part candidates (2D heatmaps position) and connection confidence (PAF~\cite{cao2018openpose} score ranging between 0 and 1) between body parts (see Fig.~\ref{fig:overview_system}(a)), together with the former reconstructed 3D skeletons. By regarding body parts and the 3D joints in the former frame as graph nodes, we construct edges with significant semantic meaning between nodes. Specifically, as shown in Fig.~\ref{fig:overview_system}(b), there exist three kinds of edges: per-view parsing edges connecting adjacent body parts in each image view, cross-view matching edges connecting the same body part across views, and temporal tracking edges connecting history 3D nodes and 2D candidates. The construction of these edges will be elaborated in Sect.~\ref{sec:construct}.

Based on the input graph in Fig.~\ref{fig:overview_system}(b), this 4D association problem can be described as a minimum-cost multi-cut problem, \textit{i.e}., a 0-1 integer programming problem to select those edges that belong to the real skeletons and the physically real temporal and cross-view edges, see Fig.~\ref{fig:overview_system}(c). Actually, our graph model is similar to the available single view association problem \cite{cao2018openpose,deepercut}, except that it is more complex. As it is a NP-hard problem, we split it to 4D limb parsing (Sect.~\ref{sec:sec:clique}) and a skeleton assembling (Sect.~\ref{sec:sec:bundle}) problems. Our proposed solving method can guarantee realtime performance while obtaining robust results. Here, it is worth mentioning that, our graph model and the solving method also work for special cases when there is no temporal edges, \textit{i.e}., at the first frame of the whole sequence, or when new persons entering the scene. 


%% file: version2/4_modeling.tex
\section{4D Association Graph}
\label{sec:construct}
For each image view $c\in\{1,2,...,N\}$ at the current frame $t$, the convolutional pose machine (CPM) model \cite{wei2016convolutional,cao2018openpose} is first applied to get the heatmaps of keypoints and their part affinity fields (PAFs).
Denote $\calD_j(c)=\{\mathbf{d}_j^m(c)\in\mathbb{R}^2\}$ as the candidate positions of the skeleton joints $j\in\{1,2,...,J\}$, with $m$ as candidate index. Here, $t$ is ignored by default as processing the current frame. Denote $f_{ij}^{mn}(c)$ as PAF score connecting $\bfd_i^m(c)$ and $\bfd_j^n(c)$, where $\{ij\}\in\calT$ is a limb on the skeleton topology tree $\calT$.

With both the candidate positions $\calD_j(c)$ and the skeleton joints reconstructed in former frame seen as graph nodes, we have three kinds of edges: per-view parsing edges $\calE_P$ connecting nodes in the same view, cross-view matching edges $\calE_V$ connecting nodes in different views geometrically, and temporal tracking edges $\calE_T$ connecting nodes temporally. The solving of this association graph is equivalent to determining bool variable $z\in\{0,1\}$ for each edge, where $z=1$ means connected nodes are associated in the same human body, $z=0$ otherwise. Note that $z=0$ means the two nodes are linked with two different bodies, or are linked with a false position (a fake joint that is not on a real body). The connecting weight on edges is written as $p(z)=p(z=1)$. In the following, the weights of each edge is defined in the 4D association graph.

\subsection{Parsing Edges and Matching Edges}
\label{sec:sec:gp_graph}
Without considering the temporal tracking edges introduced by the former reconstructed 3D skeletons, the parsing edges and the matching edges forms a 3D association graph $\calG_{3D}$. This case happens when processing the first frame of the whole sequence or when a new person is entering in the scene.
The graph $\calG_{3D}$ directly extends the original per-view multiple person parsing problem \cite{cao2018openpose} with cross view geometric matching constraints. With these geometric constraints, false limb connections in single view case may have good chance to be distinguished and corrected during joint 3D association.

Denote $z_{ij}^{mn}(c_1,c_2)$ as bool variable on edge connecting $\bfd_i^m(c_1)$ and $\bfd_j^n(c_2)$. Obviously, a feasible solution $\{z_{ij}^{mn}(c_1,c_2)\}$ on $\calG_{3D}$ must conforms to the following inequalities：
\begin{equation}\label{equ:feasible_parsing_1}
    \begin{split}
      \forall m,  & \sum_{n} {z_{ij}^{mn}(c,c)} \le 1  \\
      \forall c_2\neq c_1, m,  & \sum_{n} {z_{ii}^{mn}(c_1,c_2)} \le 1
    \end{split}
\end{equation}
Specifically, the top one forces that no two edges share a node, i.e., no two limbs of the same type (e.g., left forearm) share a part. The bottom one forces that no joint from one view connects to two joints of the same type from another view. Note also here $c_1$ and $c_2$ represent all possible combinations of view pairs.

For the per-view parsing edge $\calE_P$, we directly define the input edge weight as its PAF score:
\begin{equation}\label{equ:parsing}
p(z_{ij}^{mn}(c)=1)=f_{ij}^{mn}(c)
\end{equation}
For cross-view matching edge $\calE_V$, the weight is defined based on the epipolar distance, written as line-to-line distance in 3D space:
\begin{equation}\label{equ:cross_view_e}
    p(z_{ii}^{mn}(c_1,c_2)) = 1-\frac{1}{Z} \bfd_i^m(c_1)\oplus\bfd_i^n(c_2)
\end{equation}
\begin{equation} \label{equ:epipolar}
    \bfd(c_1)\oplus\bfd(c_2) = d(K_{c_1}^{-1}\homod(c_1), K_{c_2}^{-1}\homod(c_2))
\end{equation}
where $\homod = [\bfd^{\rmT}, 1]^{\rmT}$, $K_{c}$ is intrinsic matrix of view $c$,
$d(\cdot,\cdot)$ means line-to-line distance between two rays emitting from the camera centers of view $c_1$
and $c_2$. $Z$ is an empirically defined normalization factor, which adjusts epipolar distance to range $[0,1]$. Note that we only build edges for those cross-view nodes sharing the same joint index.

\subsection{Tracking Edges}
\label{sec:sec:tr_graph}


Although solving $\calG_{3D}$ at each time instant could provide good association in most cases, failures may happen for very crowded scene or severe occlusions. To improve skeleton reconstruction robustness, we take advantage of the temporal prior, i.e., the reconstructed skeletons at the former frame for regularization of the association problem, which forms the \textbf{4D association graph} $\calG_{4D}$. We restrict the connecting edge between the former frame skeletons and the current frame joint features, by requiring the two nodes of the edge to be the same skeleton joint (can be on different persons). Denote $z_i^{mk}(c)$ as the final optimized bool variable for edge connecting image joint feature $\bfd_i^m(c)$ and skeleton joint $\bfX_i^k$. We define tracking edge connecting probability as
\begin{equation}\label{equ:temporal}
    p(z_i^{mk}(c)) = 1-\frac{1}{T}d'(\bfX_i^k, K_{c}^{-1}\bfd_i^m(c))
\end{equation}
where $d'(\bfX, \bfd)$ indicates point-to-line distance between 3D point $\bfX$ and 3D line emitting from camera center to $\bfd$, and $T$ is normalization factor, ensuring $p(z_i^{mk}(c))$ to be in range $[0,1]$. 
Similarly, we have inequality conditions hold for the feasible solution space:
\begin{equation}\label{equ:feasible_temp}
    \begin{split}
        \forall i,c, \sum_{m} z_i^{mk}(c) \le 1, \ \sum_{k} z_i^{mk}(c) \le 1
    \end{split}
\end{equation}
This constraint forces that each 3D joint at the last frame matches no more than one 2D joint on each view at the current frame, and vice versa.
 
\subsection{Objective Function}
Based on the predefined probabilities for the parsing edges, matching edges and tracking edges, our 4D association optimization can be formulated as an edge selection problem to maximize an objective function under conditions \ref{equ:feasible_parsing_1} and \ref{equ:feasible_temp}. Specifically, let $q(z)=p(z)\cdot z$ denote the final energy of an edge, where $z$ is a boolean variable, and then our objective function can then be written as the summation of energies of all the selected edges in $\calE_P$, $\calE_M$ and $\calE_T$: 
\begin{equation}
    \label{equ:full_objective}
    \begin{split}
    E(\calZ) = & w_p \sum q(z_{ij}^{mn}(c,c)) + w_m \sum q(z_{ii}^{mn}(c_1,c_2)) \\
                 & + w_t \sum q(z_i^{mk}(c))
    \end{split}
\end{equation}
Note here $\sum$ would traverse all the possible edges, i.e., all feasible values of variables ($i$,$j$,$m$,$n$,$k$,$c$,$c1$,$c2$) by default. $w_p$, $w_m$ and $w_t$ are empirically defined weighting factors for edges $\calE_P$, $\calE_M$ and $\calE_T$, respectively. 
With $w_t=0$, it degenerates to the objective function for solving association graph $\calG_{3D}$. Notice that, both $\calG_{3D}$ and $\calG_{4D}$ can be solved with the same procedure, as described in Sect.~\ref{sec:approx}.


%% file: version2/5_solving.tex
\section{Solving 4D Association}
\label{sec:approx}

Solving the 4D Association graph means maximizing the objective function Eqn.~\ref{equ:full_objective} under constraints Eqn.~\ref{equ:feasible_parsing_1} and Eqn.~\ref{equ:feasible_temp}. Traversing the huge association space in brute force manner is infeasible for realtime systems. Instead, inspired by the realtime but high quality parsing method \cite{cao2018openpose} that assembles 2D human skeleton in a greedy manner, we propose a realtime 4D association solver. The key difference between our 4D association and the previous 2D association is that: the limb candidates scatter not only in a single image but in the whole space and time, and some limbs represent the same physical limbs. Therefore, we need to first associate those limbs that are likely to be the same limb bundle across views and times, before 4D skeletons assembling. Based on this idea, our realtime solution can be divided into two steps: 4D limb bundle association (Sect.~\ref{sec:sec:clique}), and 4D human skeleton association by the bundle Kruskal's algorithm (Sect.~\ref{sec:sec:bundle}). It is worth noting that, both of these two steps rely on the objective function Eqn.~\ref{equ:full_objective} for optimization.




\begin{figure}
  \centering
  \includegraphics[width=0.9\linewidth]{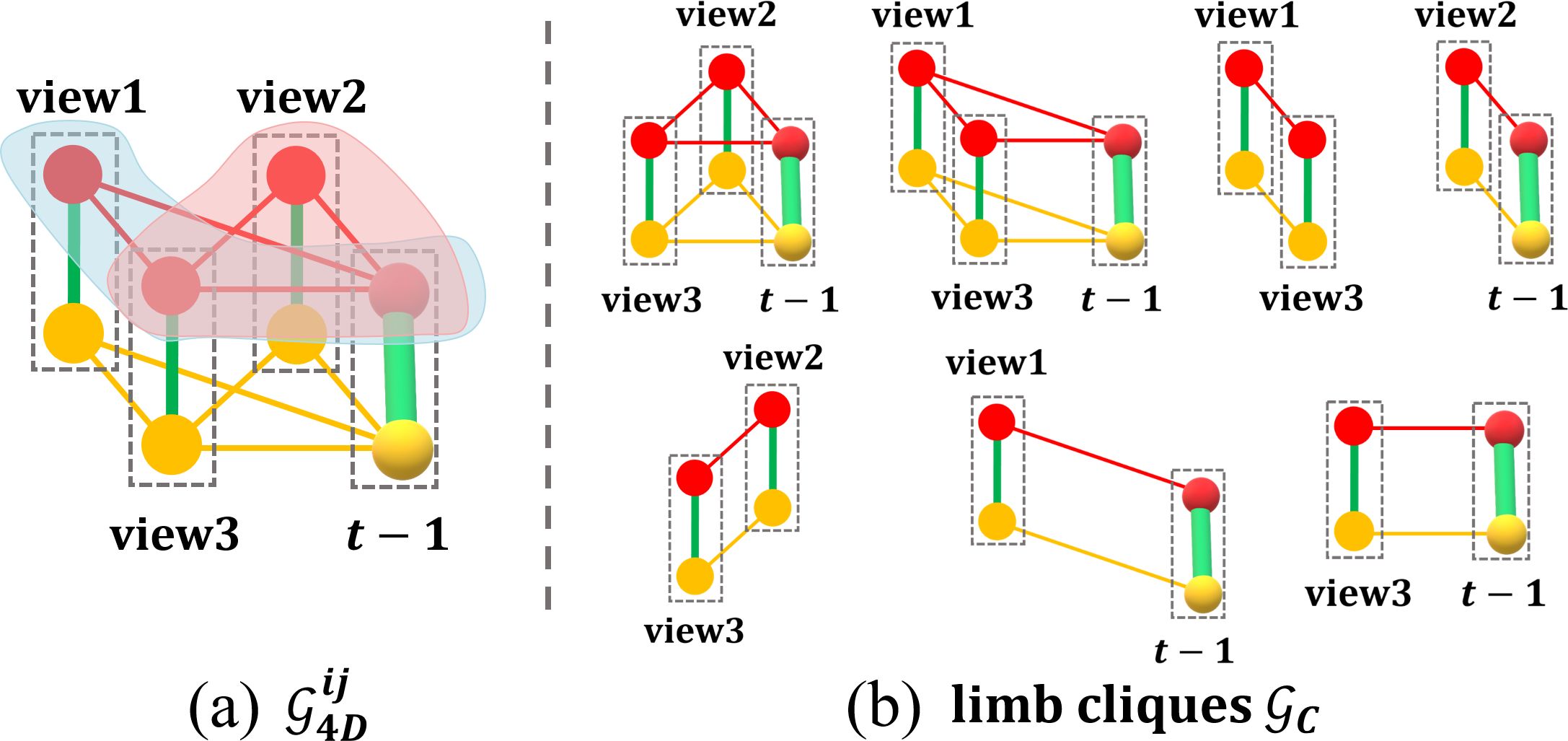}
  \caption{Illustration of limb cliques.
  (a) A sample 4D graph on limb $\{ij\}$ denoted as $\calG_{4D}^{ij}$. Two cliques are marked as red area and blue area. (b) Limb cliques of different sizes could be proposed from the 4D graph on limb. Joints of the same type (same color in the above figure) on a limb clique form a clique, and joints of different types on each view must share a green parsing edge.
  }
  \label{fig:bone_clique}
  \vspace{-3mm}
\end{figure}

\subsection{4D Limb Bundle Parsing}
\label{sec:sec:clique}

To extract limb bundles across view and time, we first restrict $\calG_{4D}$ on a limb $\{ij\}$ (two adjacent types of joint) as $\calG_{4D}^{ij}$.
Since there are multiple persons in the scene, graph $\calG_{4D}^{ij}$ may contain multiple real limb bundles. In theory, each real limb bundle contains two joint cliques. For clarity, a clique means a graph where every two nodes are connected \cite{Wilson96introductionto}, see Fig.~\ref{fig:bone_clique}(a) for example. This implies that every two joints of the same type in the limb bundle must share a cross-view edge or a temporal edge. By further considering the parsing edges, a correct 4D limb bundle consists of two joint clique connected with parsing edges on each view. We call such limb bundle candidate as \textit{limb clique}. Fig.~\ref{fig:bone_clique}(b) enumerates all the possible limb cliques of Fig.~\ref{fig:bone_clique}(a). Consequently, our goal in this step is to search
all possible limb cliques $\{\calG_C| \calG_C\subset\calG_{4D}^{ij}\}$ for the real limb bundles.

We measure each limb clique with $E(\calZ_{\calG_C})$ based on the objective function Eqn.~\ref{equ:full_objective}.
However, directly maximizing $E(\calZ_{\calG_C})$ would always encourage as many edges as possible to be selected in a clique, even false edges.
Hence, we normalize $E(\calZ_{\calG_C})$ with clique size $|\calV_C|$ of $\calG_C$, and add a penalty term to balance the clique size and the average probability.
Overall, the objective function for a limb clique is
\begin{equation} \label{equ:clique_loss}
    E(\calG_C) = E(\calZ_{\calG_C})/|\calV_C| + w_v \rho(|\calV_C|)
\end{equation}
where $w_v$ is balancing weight, and $\rho$ is a Welsch robust loss\cite{dennis1978techniques,barron2019general} defined as
\begin{equation} \label{equ:welsch}
    \rho(x) = 1-\exp\left(-\frac{1}{2}(x/c)^2\right)
\end{equation}
Here, $c = (N-1)/2$ is a parameter depending on the total number of views.

\begin{figure}
    \centering
    \includegraphics[width=0.9\linewidth]{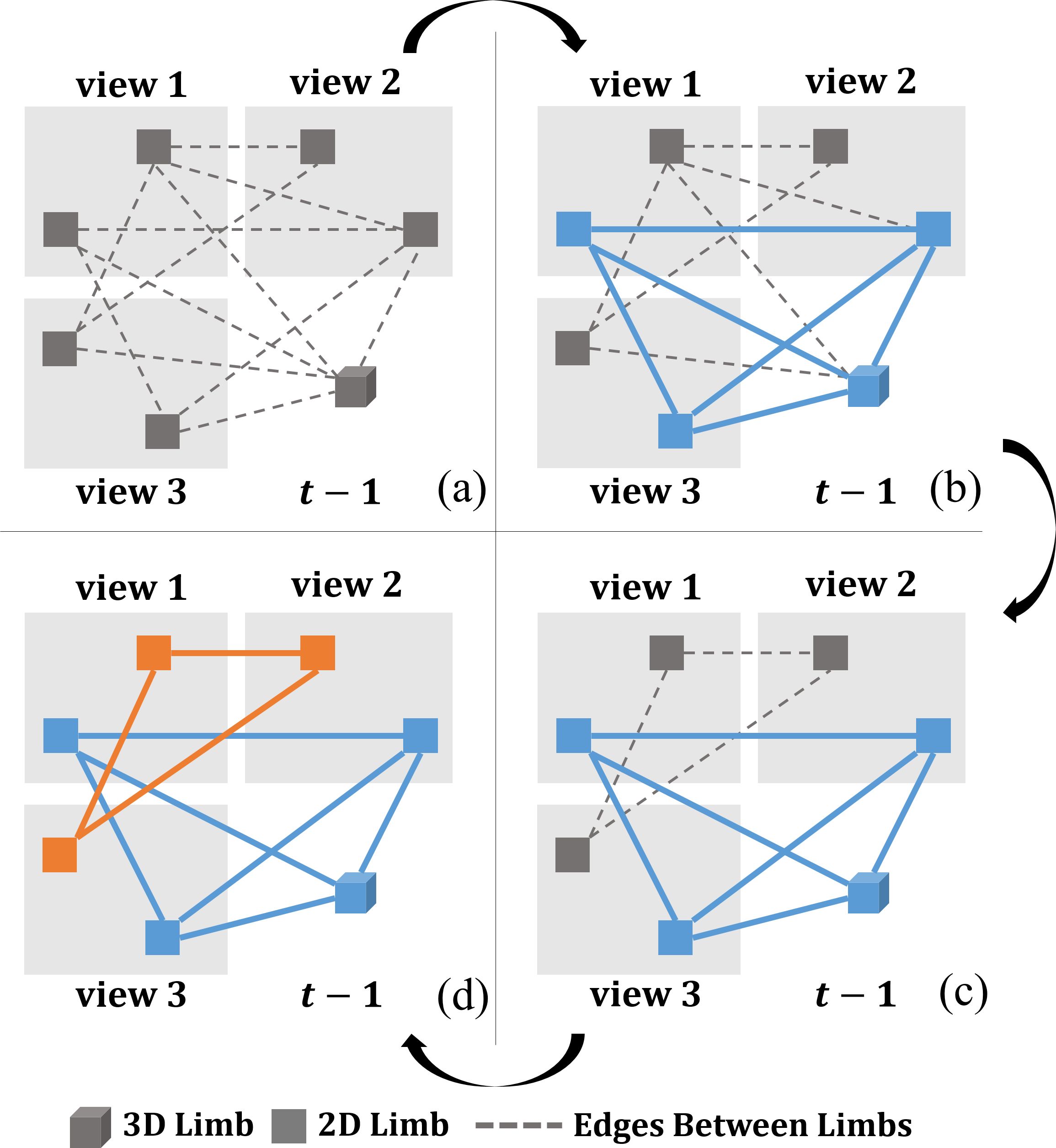}
    \caption{Illustration of limb bundle parsing procedure.
    (a) Initial graph $\calG_{4D}^{ij}$. A square/cube represents a limb (2D or 3D), and each grey dash line means an edge.
    (b) A best clique (limb bundle) detected from (a) is shown in blue.
    (c) Then, we remove both limbs and edges related to the best clique, and extract next best one.
    (d) Finally, all cliques are detected. We could extract cliques without temporal edges, like the orange one.}
    \label{fig:clique}
    \vspace{-3mm}
\end{figure}

Fig.~\ref{fig:clique} illustrates the limb bundle parsing procedure. After selecting a limb clique and marking it as a limb bundle, we remove it from $\calG_{4D}^{ij}$ (Fig.~\ref{fig:clique}(b)), together with all other edges connected with any joint in this clique (Fig.~\ref{fig:clique}(c)). By doing this, our solution always conforms to feasibility inequalities (\ref{equ:feasible_parsing_1},\ref{equ:feasible_temp}).
This selection process is iterated until $\calG_{4D}^{ij}$ is empty (Fig.~\ref{fig:clique}(d)).
%
%
%
%

\subsection{4D Skeleton Assembling}
\label{sec:sec:bundle}

\begin{figure}
    \centering
    \includegraphics[width=\linewidth]{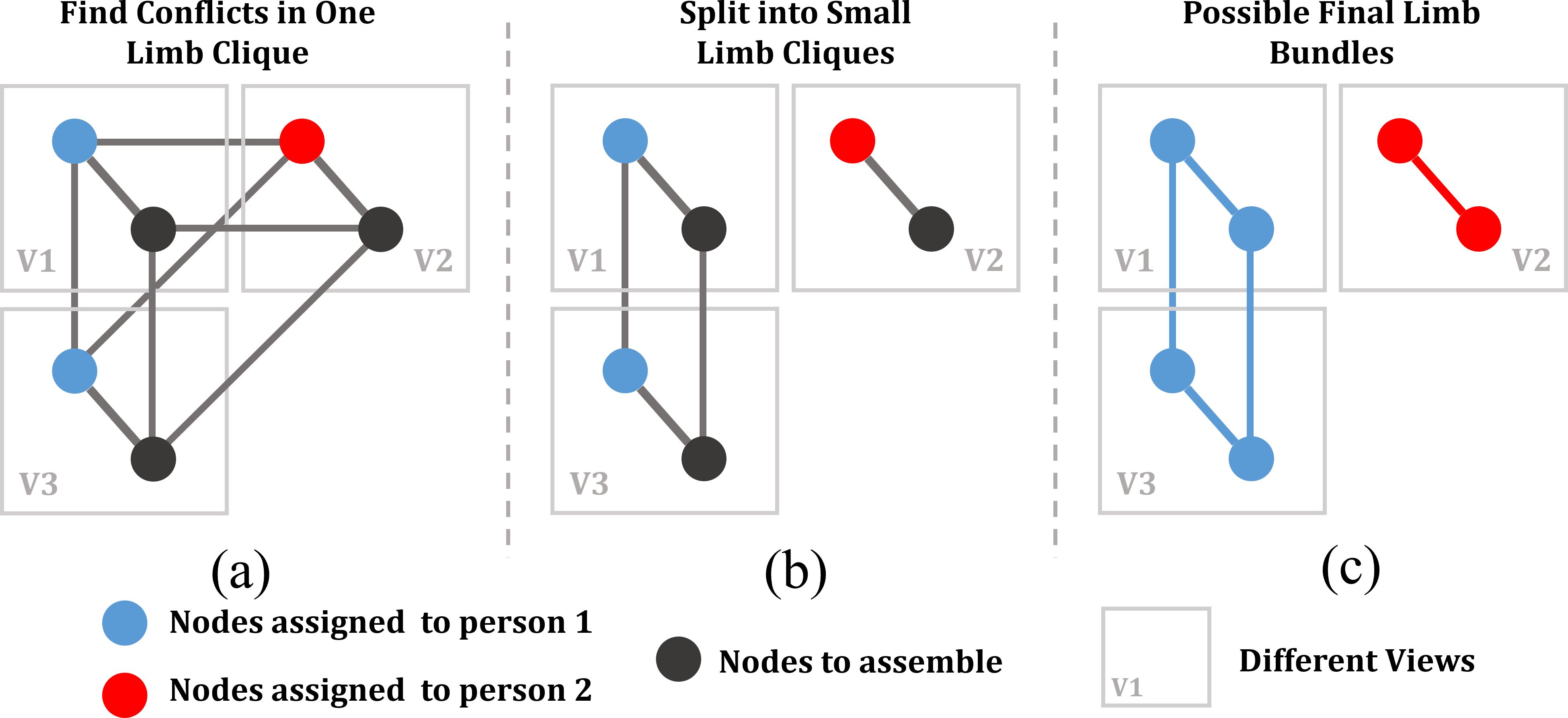}
    \vspace{-3mm}
    \caption{Conflicts handling in our skeleton assembling step.
    (a) A limb bundle to be added. It contains 3 parsing edges on 3 views. In this case, each parsing edge contains a joint to be assembled (black node) and a joint already assembled (blue or red nodes) in previous iterations.
    Here conflict arises as blue and red belong to different person IDs. (b) We split original limb bundle into small bundles according to the existing person IDs. (c) A possible final assembling result.
    }
    \label{fig:bundle}
    \vspace{-3mm}
\end{figure}

After generating all the 4D limb bundles, we need to assemble them into multiple 4D human skeletal structures. We first sort all the 4D limb bundles based on their scores, and build a priority queue to store them. In each iteration, we pop a 4D limb bundle from the queue with the maximum score (based on Eqn.~\ref{equ:clique_loss}), and merge it into the 4D skeletons. In this merging process, all the 2D joints (belongs to this bundle, from different views) should have a same labeled person ID. However, since a newly added limb bundle may share the same 4D joint as some limb bundles that are already assigned, conflicts would arise when these 2D joints have already been labeled with different person IDs on different views in the previous iterations, see Fig.~\ref{fig:bundle}(a). To eliminate this conflict, we propose a simple yet effective way by splitting the newly added limb bundles to small limb bundles according to the persons whose joints are assigned to (Fig.~\ref{fig:bundle}(b)). We then re-compute the objective function of each small bundle and push back to the prior queue for further assembling. If there is no conflict, we merge the bundle into the skeleton and label the 2D joints. We iterate popping and merging until the queue is empty (Fig.~\ref{fig:bundle}(c)).


We call the above method bundle Kruskal's algorithm. In the single view case, there would be no conflicts, and our method degenerates to traditional Kruskal's algorithm, which is a famous minimum spanning tree (MST) algorithm used in OpenPose \cite{cao2018openpose}.


\subsection{Parametric Optimization}
Based on 4D skeleton assembling results on the 2D view images, we can further optimize the full 3D body pose by embedding a parametric skeleton. We minimize the energy function
\begin{equation} \label{equ:kinematic}
    E(\Theta) = w_{2D}E_{2D} + w_{shape}E_{shape} + w_{temp}E_{temp}
\end{equation}
where $E_{2D}$ is the data term aligning 2D projections on each view to the detected joints, $E_{shape}$ penalizes human shape prior (\textit{e.g}. bone length and symmetry), and $E_{temp}$ is temporal smoothing term ($w_{2D}, w_{shape}$ and $w_{temp}$ are balancing weights, $w_{temp}=0$ if no temporal information exists). As this fitting process is a classic optimization step, please refer to \cite{bogo2016smplify,zanfir2018monocular,li2018PG} for details. Temporally, we track each person and use the average bone lengths of the first five frames with high confidence (visible in more than 3 cameras) as the bone length prior for the person in the later frames. If the person is lost and re-appear, we simply regard him/her as a new person and re-calculate the bone lengths.

%% file: version2/7_results.tex


\section{Results}
\label{sec:result}

\begin{figure*}[ht]
\begin{center}
  \includegraphics[width=0.95\linewidth]{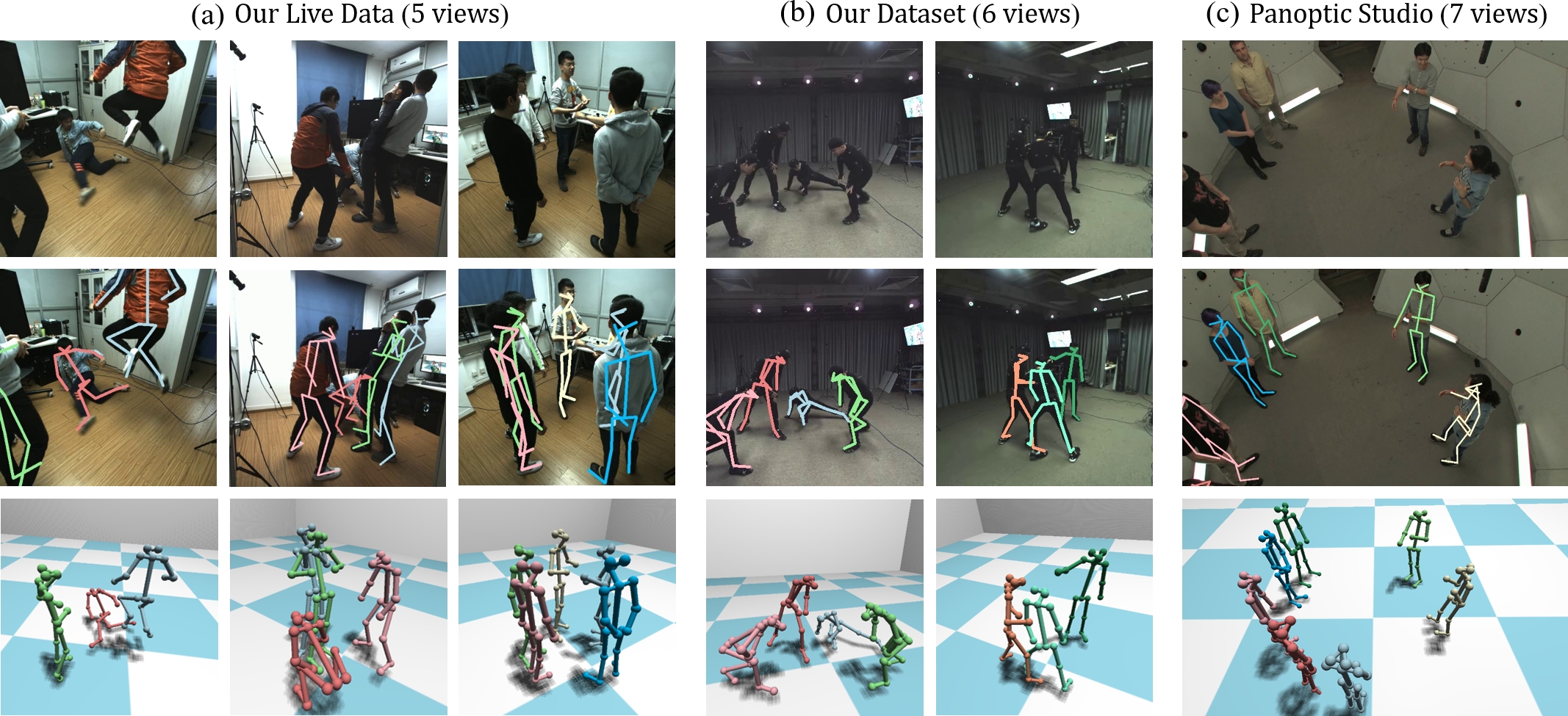}
\end{center}
  \caption{Results of our system. From top to bottom: input images, reprojection of 3D human, and 3D visualization respectively. (a) Our live captured data with fast motion (left), severe occlusion (middle) and crowded scene (right). 5 views used. (b) Our dataset with textureless clothing and rich motion. 6 views used. (c) Panoptic studio dataset with natural social interaction. 7 views used. }
\label{fig:qualitative_results}
\vspace{-3mm}
\end{figure*}

In Fig.~\ref{fig:qualitative_results}, we demonstrate the results of our system. Using only geometry information from sparse view points, our method enables realtime and robust multi-person motion capture under severe occlusions (Fig.~\ref{fig:qualitative_results}(a)), challenging poses (Fig.~\ref{fig:qualitative_results}(b)) and subtle social interactions (Fig.~\ref{fig:qualitative_results}(c)).

\subsection{Implementation Details}
\label{sec:system}

The multi-view capture system consists of 5 synchronized industrial RGB cameras (with resolution $2048\times 2048$) and a single PC with one $3.20$ GHz CPU and one NVIDIA TITAN RTX GPU. Our system achieves 30 fps motion capture for 5 persons. Specifically, for each frame, the pre-processing step (including demosaicing, undistortion and resizing for multi-view inputs) takes less than $1$ ms, the CNN inference step takes $22.9$ ms in total for $5$ images, the 4D association step takes $11$ ms, and the parametric optimization step takes less than $4$ ms. Moreover, we ping-pong the CNN inference and the 4D association for achieving realtime performance with affordable delay ($60$ ms).
More details about the optimization parameters are provided in the supplementary material.

Note that the 4D association pipeline is fully implemented on CPU. Also, in the CNN inference step, the input RGB images are resized to $368\times 368$, and the CNNs for keypoints and PAFs are re-implemented using TensorRT~\cite{vanholder2016tensorrt} for further acceleration.

\vspace{-2mm}
\subsection{Dataset}
\label{sec:dataset}
We contribute a new evaluation dataset for multi-person 3D skeleton tracking with ground truth 3D skeletons captured by commercial motion capture system, OptiTrack~\cite{optitrack}. Compared with previous 3D human datasets \cite{kazemi2013kthfootball,ionescu2013h36m,mono-3dhp2017,joo2019panoptic,belagiannis2014temporal3DPS,aa2011utrecht}, our dataset is mainly focusing on the more challenging scenarios like close interactions and challenging motion. Our dataset contains 5 sequences with each around 20-second long capturing a 2-4 person scene using 6 cameras. Our actors all wear black marker-suit for ground truth skeletal motion capture. With ground truth 3D skeletons, our dataset enables more effective quantitative evaluations for both 2D parsing and 3D tracking algorithms. Note that besides evaluating our method using the proposed dataset, we also provide evaluation results using Shelf and Panoptic Studio dataset following previous works~\cite{belagiannis2014temporal3DPS,belagiannis20163dps,dong2019fast}.





\subsection{Quantitative Comparison}
\label{sec:sec:quantitative}
We compare with state-of-the-art methods quantitatively using both the Shelf dataset and our testing dataset.
The quantitative comparison on Shelf dataset is shown in Table.~\ref{tab:sota}.
Benefiting from our 4D association formulation, we achieve more accurate results than both temporal tracking methods based on 3DPS (\cite{belagiannis2014temporal3DPS,belagiannis20143d,belagiannis20163dps,ershadi2018multiple}) and appearance-based global optimization methods \cite{dong2019fast}.

We also compare with \cite{dong2019fast} on our testing dataset according to `precision' (the ratio of correct joints in all estimated joints) and `recall' (the ratio of correct joints in all ground truth joints). A joint is correct if its Euclidean distance to ground truth joint is less than a threshold (0.2m). As shown in Table.~\ref{tab:compare_our},
our method outperforms \cite{dong2019fast} under both metrics.
\begin{table}[ht]
\begin{tabular}{lllll}
\hline
Shelf          & A1 & A2 & A3 & Avg \\ \hline
Belagiannis \etal~\cite{belagiannis20143d}  & 66.1 & 65.0 & 83.2 & 71.4 \\
$\dagger$Belagiannis \etal~\cite{belagiannis2014temporal3DPS} & 75.0 & 67.0 & 86.0 & 76.0 \\
Belagiannis \etal~\cite{belagiannis20163dps} & 75.3 & 69.7 & 87.6 & 77.5 \\
Ershadi-Nasab \etal~\cite{ershadi2018multiple} & 93.3 & 75.9 & 94.8 & 88.0 \\
Dong \etal~\cite{dong2019fast}  & 97.2   & 79.5   & 96.5            & 91.1    \\
*Dong \etal~\cite{dong2019fast}   & 98.8   & 94.1   & \textbf{97.8}   & 96.9    \\
$\dagger$\# Tanke \etal~\cite{tanke2019iterative} & \textbf{99.8} & 90.0 & 98.0 & 96.0\\
\hline
$\dagger$Ours(final)           & {99.0}   & \textbf{96.2}   & 97.6   & \textbf{97.6}    \\ \hline
\end{tabular}
\vspace{2mm}
\caption{Quantitative comparison on Shelf dataset using percentage of correct parts (PCP) metric. `*' means method with appearance information, `$\dagger$' means method with temporal information, `\#' means accuracy without head. `A1'-`A3' correspond to the results of three actors, respectively. The averaged result is in column `Avg'. }
\label{tab:sota}
\end{table}



\begin{table}[ht]
    \centering
    \begin{tabular}{llll}
    Our Dataset     &   Dong\cite{dong2019fast} & Ours(final) \\ \hline
    Precision(\%)       &  71.0 & \textbf{88.5} \\
    Recall(\%)          &  80.2 & \textbf{90.2} \\ \hline
    \end{tabular}
    \vspace{2mm}
    \caption{Comparison with \cite{dong2019fast} using our testing dataset.}
    \label{tab:compare_our}
\end{table}

\begin{figure}[ht!]
  \centering
  \includegraphics[width=\linewidth]{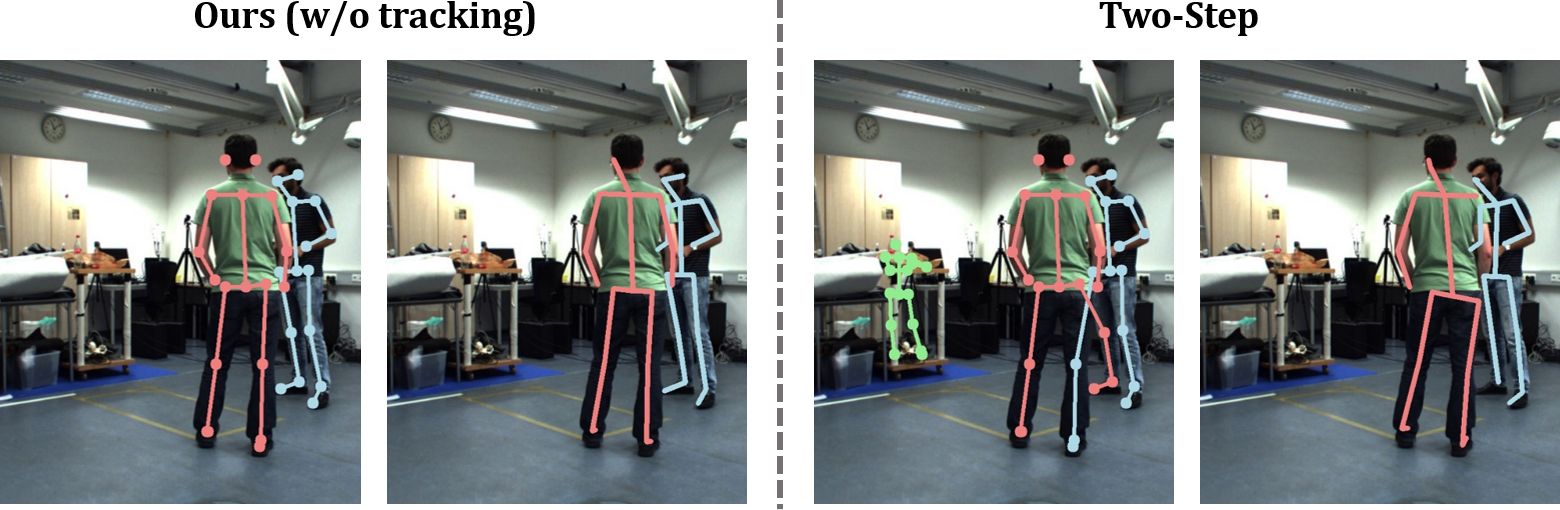}
  \caption{Comparison with two-step pipeline. Top figures are association result, bottom figures are reprojection of 3D pose. Notice that, reprojection of 3D pose generated by two-step pipeline obviously
  deviates from correct position due to false parsing.}\label{fig:two-step}
      \vspace{-3mm}
\end{figure}

\begin{figure*}[ht!]
    \centering
    \includegraphics[width=0.95\linewidth]{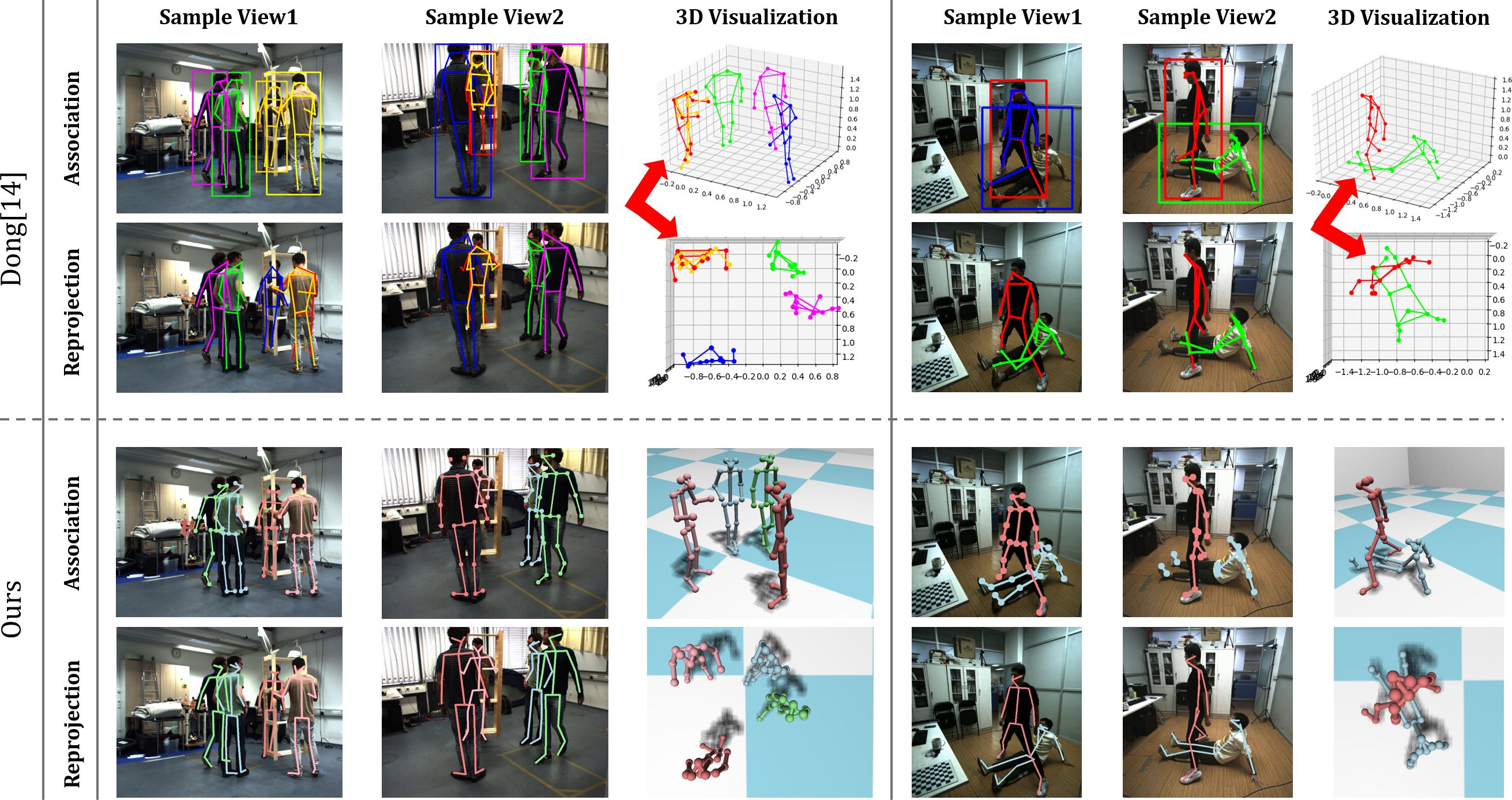}
    \caption{Qualitative comparison with Dong\cite{dong2019fast} on Shelf (left figure) and our captured data (right figure), both with 5 cameras. For each case, we show association results and reprojection of 3D pose on two sample views. For 3D visualization, we show a side view rendering and a top view rendering for clear comparison. }
    \label{fig:quali-compare}
        \vspace{-3mm}
\end{figure*}

\subsection{Qualitative Comparison}
\label{sec:sec:qualitative}
To further demonstrate the advantages of our bottom-up system, we perform qualitative comparison with the state-of-the-art method \cite{dong2019fast}, which utilizes top-down human pose detector \cite{Chen2018CPN} to perform single view parsing. The qualitative results is shown in Fig.~\ref{fig:quali-compare}, from which we can see that top-down method depends heavily on instance proposals, and may generate false positive human pose detection to deteriorate the cross-view matching performance (left case). Furthermore, per-view parsing would fail to infer correct human poses under severe occlusion, deteriorating pose reconstruction results (right). Instead, thanks to relatively precise low-level features (e.g. keypoints) and robust 4D association algorithm, the joints are associated more accurately in our results.

\subsection{Ablation Study}
\label{sec:sec:ablation}
\noindent\textbf{With/Without tracking. }
We first evaluate tracking edges in the 4D graph. By triangulating 2D bodies into 3D skeletons directly using $\calG_{3D}$, we eliminate the usage of tracking edges. The result is labeled as `w/o tracking' in Table.~\ref{tab:ablation}.
Without using tracking edges, our method still exhibits competent result and out-performs state-of-the-art method \cite{dong2019fast} (93.4\% vs 91.1\%). Moreover, our 4D association method is more robust in messy scenes (`Ours(final)' as shown in Table.~\ref{tab:ablation}).

\noindent\textbf{Compare with two-step pipeline. }
Traditional two-step pipeline would perform per-view parsing first, followed by cross-view matching. We implement a two-step pipeline for comparison, where we first parse human in each view similar to \cite{cao2018openpose}, and match them using our clique searching method with objective function defined on parsed body. Notice that no temporal information is used, and 3D poses are obtained by triangulation. Result is shown as `two-step' in Table.~\ref{tab:ablation}. As shown in Table.~\ref{tab:ablation}, our per-frame $\calG_{3D}$ solution `w/o tracking' performs better than two-step pipeline, especially on actor `A2'. To show our robustness to per-view parsing ambiguity, we use only 3 views to reconstruct 2 persons (Fig.~\ref{fig:two-step}). Wrong parsing result on one view would harm the inferred 3D pose, especially when very sparse views are available.

\begin{table}[ht]
\centering
\begin{tabular}{lllll}
\hline
Shelf             & A1        & A2         & A3         & Avg            \\ \hline
two-step       &98.1  & 83.8          & \textbf{97.6}        & 93.1           \\
w/o tracking      & 96.5      & 86.8       & 97.0       & 93.4       \\
Ours(final)       & \textbf{99.0} & \textbf{96.2} & \textbf{97.6} & \textbf{97.6}    \\ \hline
\end{tabular}
\vspace{2mm}
\caption{Ablation study on Shelf dataset. `two-step' means first per-view parsing and then cross-view matching. `w/o tracking' means we solve $\calG_{3D}$ in each frame. Both `two-step' and `w/o tracking' use triangulation to infer 3D poses. Numbers are percentage of correct parts(PCP).}
\label{tab:ablation}
\end{table}


%% file: version2/8_diss.tex
\vspace{-2mm}

\section{Conclusion}
\label{sec:discussion}
In this paper, we propose a realtime multi-person motion capture method with sparse view points. 
Build on top of the low-level detected features directly, we formulate parsing, matching and tracking problem simultaneously into a unified 4D graph association framework. 
The new 4D association formulation not only enables realtime motion capture performance, but also achieves state-of-the-art accuracy compared with other methods, especially for crowded and close interaction scenarios. Moreover, we contribute a new testing dataset for multi-person motion capture with ground truth 3D poses. 
Our system narrows the gap between laboratory markerless motion capture system and industrial applications in real world scenarios. 
Finally, our novel 4D graph formulation may stimulate future research in this topic. 